\documentclass[12pt,a4paper]{article}

\usepackage[utf8]{inputenc}
\usepackage[T1]{fontenc}
\usepackage{mathpazo}              
\usepackage{microtype}
\usepackage[margin=1in]{geometry}
\usepackage{setspace}
\usepackage{amsmath,amssymb,amsthm}
\usepackage{graphicx}
\usepackage{booktabs}
\usepackage{enumitem}
\usepackage{hyperref}
\usepackage{cleveref}
\usepackage{natbib}
\usepackage{tikz}
\usetikzlibrary{positioning, arrows.meta, decorations.pathmorphing, calc}
\usepackage{epigraph}
\usepackage{xcolor}
\usepackage{caption}

\hypersetup{
  colorlinks=true,
  linkcolor=blue!60!black,
  citecolor=blue!60!black,
  urlcolor=blue!60!black
}

\title{%
  \textbf{The Non-Optimality
  of Scientific Knowledge:}\\[6pt]
  \Large Path Dependence, Lock-In, and The Local Minimum Trap
}

\author{%
  \textsc{Mohamed Mabrok}\\
  \textit{Qatar University}\\
  \texttt{m.a.mabrok@gmail.com}
}
\date{}

\begin{document}

\maketitle

\begin{abstract}
\noindent
Science is widely regarded as humanity's most reliable method for uncovering truths about the natural world. Yet the \emph{trajectory} of scientific discovery is rarely examined as an optimization problem in its own right. This paper argues that the body of scientific knowledge, at any given historical moment, represents a \emph{local optimum} rather than a global one--that the frameworks, formalisms, and paradigms through which we understand nature are substantially shaped by historical contingency, cognitive path dependence, and institutional lock-in. Drawing an analogy to gradient descent in machine learning, we propose that science follows the steepest local gradient of tractability, empirical accessibility, and institutional reward, and in doing so may bypass fundamentally superior descriptions of nature. We develop this thesis through detailed case studies spanning mathematics, physics, chemistry, biology, neuroscience, and statistical methodology. We identify three interlocking mechanisms of lock-in--cognitive, formal, and institutional--and argue that recognizing these mechanisms is a prerequisite for designing meta-scientific strategies capable of escaping local optima. We conclude by proposing concrete interventions and discussing the epistemological implications of our thesis for the philosophy of science.

\end{abstract}

\newpage
\tableofcontents
\newpage

\section{Introduction: Science as Optimization}
\label{sec:intro}

\epigraph{The greatest enemy of knowledge is not ignorance, it is the illusion of knowledge.}{--attributed to Daniel J.\ Boorstin}

The modern scientific enterprise is a monument to human ingenuity. From the Standard Model of particle physics to the sequencing of the human genome, our accumulated knowledge is breathtaking in scope and predictive power. Yet a subtle and unsettling question lurks beneath these achievements: \emph{Is our current scientific knowledge the best possible knowledge we could have obtained, given the same amount of collective effort?}

This paper argues that the answer is almost certainly no.

We propose that the development of scientific knowledge can be usefully modeled as an optimization process over a rugged landscape of possible theories, formalisms, and paradigms. Like gradient descent in machine learning--the iterative algorithm that adjusts parameters by following the steepest downward slope to minimize a loss function--science tends to follow the direction of greatest local improvement. Researchers build on existing frameworks, extend proven methods, and solve the next tractable problem. This process is enormously productive, but it carries a fundamental limitation: \emph{gradient descent converges to local minima, not necessarily global ones.}

\begin{figure}[h]
\centering
\begin{tikzpicture}[scale=1.0]
  \draw[thick, blue!70!black, smooth] plot[domain=0:12, samples=200] 
    (\x, {2.5 + 1.2*sin(60*\x) + 0.7*cos(110*\x) + 0.4*sin(200*\x) - 0.08*(\x-6)^2 + 0.004*(\x-6)^3});
  
  \filldraw[red!70!black] (3.15, 1.65) circle (3pt);
  \node[below, red!70!black, font=\footnotesize\bfseries] at (3.15, 1.45) {Current science};
  \node[below, red!70!black, font=\footnotesize] at (3.15, 1.05) {(local minimum)};
  
  \filldraw[green!50!black] (9.5, 0.6) circle (3pt);
  \node[below, green!50!black, font=\footnotesize\bfseries] at (9.5, 0.4) {Optimal framework};
  \node[below, green!50!black, font=\footnotesize] at (9.5, 0.0) {(global minimum)};
  
  \draw[<->, dashed, thick, orange!70!black] (4.5, 1.8) -- (4.5, 3.5);
  \node[right, orange!70!black, font=\footnotesize] at (4.55, 2.65) {Energy barrier};
  \node[right, orange!70!black, font=\footnotesize] at (4.55, 2.25) {(paradigm shift cost)};
  
  \draw[->, thick] (-0.3, 0) -- (12.5, 0) node[right, font=\small] {Framework space};
  \draw[->, thick] (0, -0.3) -- (0, 5) node[above, font=\small] {Explanatory cost};
\end{tikzpicture}
\caption{The scientific landscape as a rugged optimization surface. Science follows the local gradient (analogous to gradient descent) and converges to the nearest minimum. A globally superior framework may exist but be separated by a high energy barrier--the cognitive, institutional, and formal costs of paradigm transition.}
\label{fig:landscape}
\end{figure}
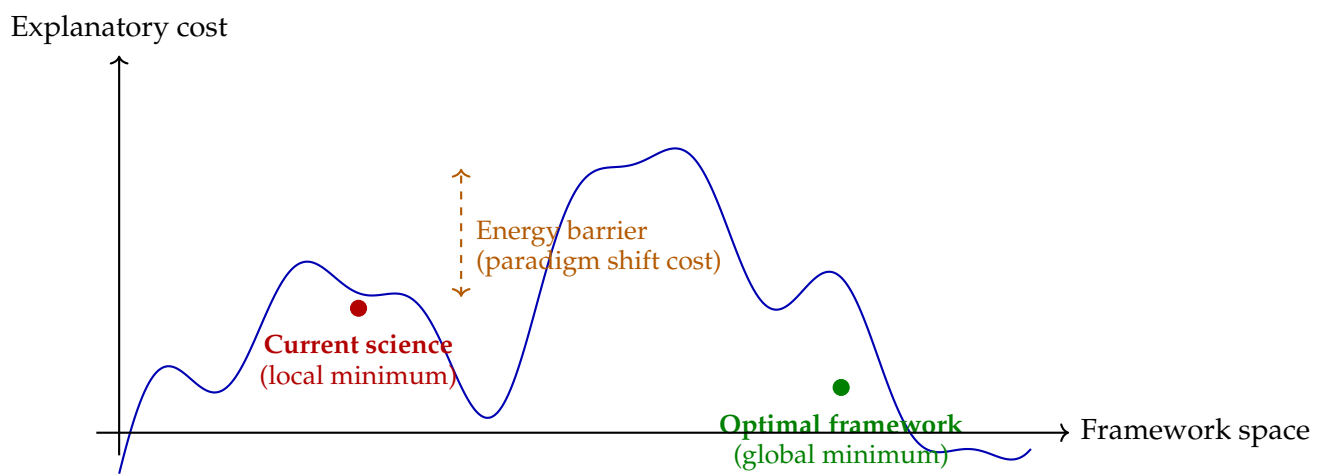

In machine learning, various strategies exist for escaping local minima: simulated annealing (introducing controlled randomness), momentum-based methods (which allow the optimizer to ``overshoot'' shallow minima), and population-based approaches like evolutionary algorithms. We argue that science has partial analogues to some of these strategies--Kuhn's paradigm shifts resemble large random jumps, for instance--but lacks systematic mechanisms for ensuring that the global landscape is adequately explored.

The structure of this paper is as follows. In \Cref{sec:framework}, we develop the formal analogy between scientific progress and optimization. In \Cref{sec:mechanisms}, we identify three interlocking mechanisms of lock-in: cognitive, formal, and institutional. In \Cref{sec:cases}, we present detailed case studies demonstrating non-optimality across multiple scientific domains. In \Cref{sec:counterarguments}, we address objections to our thesis. In \Cref{sec:escape}, we propose concrete strategies for escaping local optima. Finally, in \Cref{sec:implications}, we discuss the broader epistemological implications.

\section{A Framework for Scientific Non-Optimality}
\label{sec:framework}

\subsection{The Optimization Analogy}

Let us formalize the intuition. Define $\mathcal{F}$ as the space of all possible scientific frameworks--every conceivable formalism, notation, conceptual vocabulary, and methodological approach that could be used to describe natural phenomena. Each framework $f \in \mathcal{F}$ has an associated \emph{explanatory cost} $C(f)$, which we can think of as a composite measure of:

\begin{enumerate}[label=(\roman*)]
  \item \textit{Predictive residual}: phenomena that the framework cannot explain or predict.
  \item \textit{Computational complexity}: the difficulty of extracting predictions from the framework.
  \item \textit{Conceptual opacity}: the degree to which the framework obscures rather than illuminates the underlying structure of nature.
\end{enumerate}

An \emph{optimal} scientific framework $f^*$ minimizes $C(f)$ globally:
\begin{equation}
  f^* = \arg\min_{f \in \mathcal{F}} \; C(f).
  \label{eq:global}
\end{equation}

Science, as historically practiced, does not perform this global optimization. Instead, it performs something much closer to iterative local search:
\begin{equation}
  f_{t+1} = f_t - \eta \, \nabla_{f} C(f)\big|_{f=f_t}
  \label{eq:gradient}
\end{equation}
where $\eta$ represents the ``learning rate'' of the scientific community--how quickly it can assimilate incremental changes--and $\nabla_f C$ represents the local gradient of improvement accessible from the current position.

\subsection{Properties of the Scientific Landscape}

The analogy becomes powerful when we recognize that $\mathcal{F}$ is almost certainly a \emph{rugged landscape} in the sense of Kauffman's $NK$ models \citep{kauffman1993origins}. Several properties make this plausible:

\medskip\noindent\textit{High dimensionality.} The space of possible frameworks is astronomically large. A scientific framework involves choices of formalism (differential equations vs.\ algebraic structures vs.\ computational models), ontological commitments (what entities are taken as fundamental), mathematical infrastructure (number systems, geometries, logics), and methodological principles (reductionism vs.\ holism, perturbative vs.\ non-perturbative approaches).

\medskip\noindent\textit{Epistatic interactions.} Choices in one dimension constrain choices in others. Adopting calculus as a primary formalism, for example, privileges continuous models over discrete ones, smooth functions over fractal structures, and local descriptions over global ones. These interdependencies create the ``ruggedness'' that generates multiple local optima.

\medskip\noindent\textit{Variable basin widths.} Some local minima have very wide basins of attraction--Newtonian mechanics, for instance, is so broadly useful that small perturbations always lead back to it. This makes escape particularly difficult.

\subsection{The Path Dependence Thesis}

Our central claim can now be stated precisely:

\begin{quote}
\textbf{Thesis (The Local Minimum Trap).} \emph{The current body of scientific knowledge, $f_{\text{now}}$, satisfies $\nabla_f C(f)\big|_{f_{\text{now}}} \approx 0$, meaning that no small modifications to existing frameworks yield significant improvement, while $C(f_{\text{now}}) > C(f^*)$, meaning that fundamentally different frameworks with lower explanatory cost exist but are inaccessible via incremental refinement.}
\end{quote}

This is a claim about path dependence: the specific trajectory of scientific history--which problems were tackled first, which tools were available, which cultural and cognitive biases prevailed--has channeled us into a particular basin of attraction from which escape requires discontinuous jumps.

\section{Three Mechanisms of Lock-In}
\label{sec:mechanisms}

We identify three interlocking mechanisms through which science becomes trapped in local optima.

\subsection{Cognitive Lock-In}

Human cognition is not a neutral substrate for scientific reasoning. Our evolved cognitive architecture introduces systematic biases into the frameworks we construct:

\medskip\noindent\textit{Linearization bias.} Human intuition handles linear relationships far more easily than nonlinear ones. This bias has shaped the development of physics from Newton onward: we linearize problems wherever possible, treat nonlinearities as perturbations, and classify systems as ``complex'' or ``intractable'' precisely when they resist linearization. But nature is fundamentally nonlinear. The perceived difficulty of turbulence, protein folding, and consciousness may partly reflect the mismatch between our linear cognitive templates and the nonlinear structure of reality.

\medskip\noindent\textit{Spatial and visual bias.} Humans are intensely visual and spatial thinkers. This has privileged geometric and spatial formalisms in physics (phase spaces, manifolds, fiber bundles) while relatively neglecting algebraic, combinatorial, and information-theoretic descriptions. When a physicist says a result is ``intuitive,'' they almost always mean it can be visualized spatially. This is a feature of human cognition, not of nature.

\medskip\noindent\textit{Reductionist bias.} Our cognitive architecture finds it natural to decompose wholes into parts and to seek explanations at the level of elementary constituents. While reductionism has been spectacularly successful, it may systematically undervalue emergent, holistic, and relational descriptions. The persistent difficulty of understanding consciousness, ecosystems, and economic systems may reflect not their intrinsic complexity but the limitations of a reductionist lens.

\medskip\noindent\textit{Narrative and causal bias.} Humans are story-telling animals. We instinctively seek causes, construct narratives, and explain phenomena in terms of mechanisms. This biases us toward mechanical, causal models and away from acausal descriptions (statistical patterns, symmetry principles, information-theoretic constraints) that might be more fundamental.

\subsection{Formal Lock-In}

The mathematical and notational infrastructure of science creates its own form of path dependence.

\medskip\noindent\textit{The calculus monopoly.} Modern science is built on the calculus of Newton and Leibniz. Differential equations are the default language for describing change, and the analytical techniques developed over three centuries constitute an enormous body of craft knowledge. But calculus was designed for smooth, continuous functions. When nature is discrete, stochastic, or fractal, we force it into continuous approximations--not because continuity is true, but because our tools demand it.

\medskip\noindent\textit{Notational inertia.} Mathematical notation shapes thought in ways that are difficult to perceive from within a tradition. Leibniz's notation for calculus ($\frac{dy}{dx}$) encourages certain manipulations (treating derivatives as fractions) while obscuring others (the distinction between partial and total derivatives, the role of the chain rule in coordinate transformations). Dirac's bra-ket notation revolutionized quantum mechanics partly by making certain computations \emph{notationally natural}. How many insights remain hidden behind awkward notation?

\medskip\noindent\textit{Foundational commitments.} The choice of mathematical foundations--set theory vs.\ category theory vs.\ type theory vs.\ homotopy type theory--shapes what can be easily expressed and what remains tortured. Much of modern mathematics is written in the language of Zermelo-Fraenkel set theory, but category theorists have argued persuasively that many structures are more naturally described categorically. The difficulty of certain mathematical problems may be artifacts of foundational choices rather than intrinsic features.

\subsection{Institutional Lock-In}

The social and economic organization of science reinforces cognitive and formal lock-in through powerful feedback loops.

\medskip\noindent\textit{The publication ecosystem.} Peer review evaluates novelty and rigor \emph{within} existing paradigms. A paper that proposes a fundamentally new formalism for fluid dynamics must overcome the skepticism of reviewers trained in the old formalism. The result is an asymmetry: incremental advances within the current paradigm face low barriers to publication, while radical alternatives face very high ones.

\medskip\noindent\textit{Educational pipelines.} Graduate training is, by necessity, an apprenticeship in existing methods. A physics Ph.D.\ student spends years mastering perturbation theory, Feynman diagrams, and differential geometry. This investment creates cognitive switching costs: even if a superior formalism exists, the individual researcher has strong incentives to continue working within the framework they have already mastered.

\medskip\noindent\textit{Funding structures.} Grant agencies evaluate proposals partly on the basis of preliminary results and methodological feasibility. Work within established paradigms can demonstrate feasibility far more easily than work that proposes to rebuild foundational methods. The result is a systematic bias toward incremental science.

\medskip\noindent\textit{Prestige economies.} Scientific prestige accrues to those who solve problems within recognized frameworks. The Fields Medal, the Nobel Prize, and named professorships all reward mastery of existing paradigms. A researcher who spends a career developing a radically new formalism that does not yet solve any recognized problem receives little institutional reward, even if the formalism might eventually prove superior.

These three mechanisms--cognitive, formal, and institutional--are mutually reinforcing. Cognitive biases shape the formalisms we develop; formalisms constrain the questions we can ask; institutional structures reward work within existing formalisms and penalize departures. The result is a powerful basin of attraction that resists escape.

\subsection{Sociopolitical Lock-In: War, Power, and the Direction of Science}
\label{sec:sociopolitical}

Beyond cognitive, formal, and institutional mechanisms, there exists a fourth and frequently underestimated form of lock-in: the sociopolitical forces--wars, geopolitical rivalries, colonial ambitions, and ideological competitions--that channel scientific inquiry along trajectories determined not by epistemic optimality but by the interests of dominant powers.

\medskip\noindent\textit{The aerodynamics wars.} The history of aerodynamic theory provides a striking illustration. In the early twentieth century, two rival schools competed to develop a theory of lift: the Cambridge school of mathematical physics (Rayleigh, Lamb, Taylor, Bairstow) and the G\"{o}ttingen school of technical mechanics (Prandtl, Kutta, Munk, Betz). The intellectual contest was inseparable from the Anglo-German military rivalry. Frederick Lanchester, a British engineer, proposed a circulation theory of lift as early as 1894, but his ideas were dismissed by the Cambridge establishment--partly because he was an outsider without elite academic credentials, and partly because the British school was ideologically committed to the mathematical rigor of exact solutions over the pragmatic approximations favored by the Germans \citep{bloor2011enigma}. The Germans, trained in the tradition of \emph{Technische Mechanik}, embraced Prandtl's boundary layer theory and Kutta's circulation condition, which--despite being based on idealized inviscid flow--produced workable engineering predictions.

When the Allied victory in World War I and subsequently World War II established Anglo-American scientific hegemony, the theoretical frameworks that had proven useful for wartime aircraft design became the standard curriculum worldwide. The Navier-Stokes equations and Kutta's condition were not adopted because they were the optimal description of fluid dynamics; they were adopted because the nations that used them built the aircraft that won the wars. As historian David Bloor documented, the British eventually accepted the German circulation theory they had initially rejected--not because new evidence settled the debate, but because the theory worked well enough and the institutional momentum of wartime aeronautics made alternatives invisible.

\medskip\noindent\textit{The Manhattan Project and the nuclear channeling of physics.} The Manhattan Project represents perhaps the most dramatic example of war reshaping the trajectory of science. The urgency of developing nuclear weapons during World War II redirected an entire generation of physicists--Oppenheimer, Fermi, Bethe, Feynman, Teller, and hundreds of others--from fundamental research into weapons engineering. The postwar scientific landscape was permanently altered: national laboratories built for weapons research (Los Alamos, Lawrence Livermore, Oak Ridge) became major centers of physics research, and the questions they asked were shaped by their military origins. Nuclear physics and particle physics received enormous funding; other areas of physics--condensed matter, fluid dynamics, nonlinear dynamics--were comparatively starved. The ``big science'' model born from the Manhattan Project and the MIT Radiation Laboratory privileged large-scale, expensive, reductionist programs and marginalized small-scale, exploratory, and cross-disciplinary work.

\medskip\noindent\textit{The Cold War and the Space Race.} The Cold War competition between the United States and the Soviet Union channeled vast resources into specific technological trajectories: rocketry, satellite technology, nuclear weapons, and computing. These investments produced extraordinary achievements, but they also created powerful path dependencies. The chemical rocket paradigm for space launch, for instance, became so deeply entrenched--through infrastructure, expertise, industrial base, and institutional commitment--that alternative propulsion concepts (nuclear thermal, ion drives, solar sails) were marginalized for decades despite their theoretical superiority for certain mission profiles. More broadly, the Cold War framed science as a tool of national prestige and military power, systematically privileging research with defense applications over basic science motivated by curiosity alone.

\medskip\noindent\textit{Colonial science and epistemic hegemony.} The globalization of Western science through colonialism and post-colonial institutional structures has created a more subtle form of sociopolitical lock-in. The scientific frameworks developed in Western Europe--Newtonian mechanics, Cartesian reductionism, frequentist statistics--were exported worldwide through colonial education systems and became the universal language of science. Alternative intellectual traditions--Chinese algebraic approaches, Indian mathematical frameworks, Islamic contributions to optics and astronomy--were absorbed selectively or marginalized entirely. The result is a global scientific enterprise operating within a remarkably narrow band of the conceptual landscape, with the illusion of universality masking deep historical contingency.

These four mechanisms--cognitive, formal, institutional, and sociopolitical--form a self-reinforcing system. Wars determine which nations dominate; dominant nations export their scientific frameworks; those frameworks shape education, funding, and institutional structures; and those structures constrain cognition and formalism. The basin of attraction deepens with each reinforcing cycle.

\section{Case Studies in Scientific Non-Optimality}
\label{sec:cases}

\subsection{Differential Equations and the Modeling of Dynamical Systems}
\label{sec:case-de}

Our first and most detailed case study concerns the role of differential equations in the physical sciences.

The Newtonian paradigm describes the evolution of physical systems through ordinary and partial differential equations. This framework is so deeply embedded in scientific practice that it is rarely questioned as a \emph{choice}. Yet it embodies specific commitments: continuous time, smooth state spaces, local interactions, and deterministic evolution (with stochastic extensions treated as generalizations rather than fundamentals).

The difficulties encountered in fluid dynamics provide a striking illustration. The Navier-Stokes equations, which describe the motion of viscous fluids, have been known since the 1840s. Yet:

\begin{itemize}[leftmargin=2em]
  \item The existence and smoothness of solutions in three dimensions remains one of the seven Millennium Prize Problems--a \$1 million bounty for a proof that has resisted nearly two centuries of effort.
  \item Turbulence, the most common state of fluid flow in nature, lacks a satisfactory analytical theory. As Werner Heisenberg reportedly quipped, turbulence is the last unsolved problem of classical physics.
  \item Computational fluid dynamics, despite enormous advances, still struggles with the vast range of scales in turbulent flows. Direct numerical simulation of turbulence at realistic Reynolds numbers remains computationally infeasible.
\end{itemize}

The conventional interpretation is that turbulence is simply a very hard problem. Our thesis suggests an alternative: \emph{the difficulty may lie partly in the formalism}. The Navier-Stokes equations describe fluid motion through velocity and pressure fields evolving under continuous partial differential operators. But real fluids are made of discrete molecules. The continuum description is an approximation, and it is precisely the approximation that may be the source of the mathematical intractability.

Evidence for this interpretation comes from alternative approaches. Lattice Boltzmann methods, which simulate fluid flow through discrete particle interactions on a lattice, often capture complex flow phenomena--multiphase flows, porous media flows, complex geometries--more naturally than PDE-based methods \citep{succi2018lattice}. Cellular automata models, pioneered by Frisch, Hasslacher, and Pomeau \citep{frisch1986lattice}, demonstrated that hydrodynamic behavior can emerge from extremely simple discrete rules without any reference to the Navier-Stokes equations.

More provocatively, the framework of computational mechanics developed by Crutchfield and others \citep{crutchfield2012between} suggests that dynamical systems may be more naturally described in terms of their computational structure--the patterns of information processing they perform--than through their differential equations. If this is correct, then the ``hardness'' of turbulence is partly an artifact of asking the wrong question in the wrong language.

\subsection{Chemistry: The Molecular Paradigm}
\label{sec:case-chem}

Modern chemistry is organized around discrete molecules connected by bonds--a conceptual framework descending from Dalton's atomic theory and Lewis's electron-pair model. This framework is extraordinarily useful for organic chemistry, drug design, and materials science. But quantum mechanics reveals it to be a convenient fiction.

In quantum chemistry, the fundamental description is a many-electron wavefunction $\Psi(\mathbf{r}_1, \mathbf{r}_2, \ldots, \mathbf{r}_N)$ (or, equivalently, the electron density $\rho(\mathbf{r})$ in density functional theory). There are no ``bonds'' in this description--only a continuous electron density distribution. The concept of a chemical bond is an interpretation imposed on the quantum-mechanical reality, not a feature of it.

This matters because enormous computational effort is devoted to translating between the quantum description and the molecular-bond description. The entire apparatus of molecular orbital theory--bonding and antibonding orbitals, hybridization, resonance structures--is a set of tools for approximately recovering the classical molecular picture from quantum mechanics. But why should this be the goal?

Alternative frameworks exist. The quantum theory of atoms in molecules (QTAIM), developed by Bader \citep{bader1990atoms}, describes chemical structure entirely in terms of the topology of the electron density, without reference to orbitals or bonds. Information-theoretic approaches characterize chemical bonding through Shannon entropy and mutual information of electron distributions. These approaches sometimes provide cleaner descriptions of phenomena (such as metallic bonding, delocalized aromatic systems, or exotic bonding in heavy-element compounds) that the traditional molecular framework handles awkwardly.

The persistence of the bond-centric paradigm is a textbook example of lock-in: textbooks, software packages (Gaussian, ORCA, VASP), visualization tools, and the entire intuitive vocabulary of working chemists are built around bonds and molecular orbitals. Switching costs are astronomical.

\subsection{The Gene-Centric View of Biology}
\label{sec:case-bio}

The Central Dogma of molecular biology--DNA makes RNA makes protein--has organized biological research for over six decades. The gene is treated as the fundamental unit of heredity, function, and evolutionary selection. This framework drove the Human Genome Project, underpins modern genomics, and shapes funding priorities across the biological sciences.

Yet post-genomic biology has revealed a far more complex picture:

\medskip\noindent\textit{Epigenetics.} Gene expression is regulated by chemical modifications to DNA and histones that do not alter the sequence itself. These modifications can be heritable across cell divisions and, in some cases, across generations. The ``code'' is not just in the sequence.

\medskip\noindent\textit{Gene regulatory networks.} The behavior of a cell depends not on individual genes but on the complex network of interactions among genes, regulatory elements, and signaling pathways. The same genome produces neurons and liver cells; the difference lies in the regulatory architecture.

\medskip\noindent\textit{Non-coding RNA.} The vast majority of the genome does not code for proteins. Much of it is transcribed into RNA molecules with regulatory functions that are only beginning to be understood. The gene-centric view, which identifies function with protein-coding sequences, systematically undervalues this enormous regulatory landscape.

\medskip\noindent\textit{The holobiont.} Organisms are not autonomous genetic entities but communities of host and microbial genomes--holobionts. The human microbiome contains more genes than the human genome itself, and these microbial genes contribute to digestion, immunity, and even behavior.

An alternative, systems-level description of biology--one that takes networks, information flow, and multi-scale organization as primary--might prove more natural than the gene-centric view. Theoretical biologists like Stuart Kauffman \citep{kauffman1993origins} and Denis Noble \citep{noble2006music} have argued precisely this. But the institutional infrastructure of biology--genome databases, sequencing pipelines, gene-centric model organisms--creates enormous inertia.

\subsection{Physics: The Perturbative Trap}
\label{sec:case-phys}

The most powerful analytical tool in theoretical physics is perturbation theory: solving problems by starting from a known, exactly solvable system and treating additional interactions as small corrections. This approach underlies quantum electrodynamics (the most precisely tested theory in physics), much of condensed matter theory, and the standard approach to quantum field theory via Feynman diagrams.

But perturbation theory \emph{requires} a small parameter. When interactions are strong--when there is no small parameter--the entire apparatus collapses:

\medskip\noindent\textit{Quark confinement.} Quantum chromodynamics (QCD) describes the strong nuclear force, but at the energy scales relevant to the binding of quarks into protons and neutrons, the coupling constant is of order unity. Perturbation theory is useless. After decades of effort, quark confinement remains analytically unproven; our best evidence comes from brute-force lattice QCD simulations.

\medskip\noindent\textit{High-temperature superconductivity.} Discovered in 1986, high-$T_c$ superconductivity remains without a satisfactory theoretical explanation. The strong electron-electron correlations in these materials defeat perturbative approaches. The Hubbard model, believed to capture the essential physics, cannot be solved analytically in the relevant parameter regime.

\medskip\noindent\textit{Quantum gravity.} The incompatibility of general relativity and quantum mechanics manifests most acutely when gravitational interactions are strong (near singularities, at the Planck scale). Perturbative quantum gravity is non-renormalizable, suggesting that entirely new non-perturbative frameworks are needed.

The AdS/CFT correspondence (gauge/gravity duality), discovered by Maldacena \citep{maldacena1999large}, provides a striking example of what escape from the perturbative trap looks like. By mapping strongly coupled gauge theories to weakly coupled gravitational theories in a higher-dimensional space, it transforms intractable strong-coupling problems into solvable weak-coupling problems. The conceptual leap required--that a theory of gravity in five dimensions is secretly the same as a gauge theory in four dimensions--is precisely the kind of discontinuous jump across the landscape that our thesis predicts is needed.

\subsection{Neuroscience: The Neuron Doctrine}
\label{sec:case-neuro}

The neuron doctrine--the principle that the neuron is the fundamental structural and functional unit of the nervous system--has guided neuroscience since Ram\'{o}n y Cajal's histological studies in the late 19th century. Modern computational neuroscience models the brain as a network of point-like neurons connected by weighted synapses, with learning implemented as changes in synaptic weights.

This framework has been enormously productive, but it may constitute a local minimum:

\medskip\noindent\textit{Glial computation.} Astrocytes, oligodendrocytes, and other glial cells outnumber neurons in the human brain and form their own signaling networks. There is growing evidence that glia modulate synaptic transmission, synchronize neural activity, and participate in information processing in ways not captured by neuron-only models \citep{fields2009other}.

\medskip\noindent\textit{Dendritic computation.} Individual dendrites perform complex nonlinear computations--including direction selectivity, coincidence detection, and multiplexing--that are not captured by point-neuron models. A single biological neuron may be computationally equivalent to a multi-layer artificial neural network \citep{beniaguev2021single}.

\medskip\noindent\textit{Electromagnetic field effects.} The brain's electromagnetic field, generated by collective neural activity, may itself carry information and influence neural dynamics (the ``field theory of consciousness'' proposed by McFadden \citep{mcfadden2020integrating}). Current models treat the field as an epiphenomenon.

\medskip\noindent\textit{Non-synaptic communication.} Neurons communicate not only through synapses but through gap junctions, volume transmission of neurotransmitters, and extracellular vesicles. These channels are largely absent from standard models.

If these neglected factors are computationally significant, then our current models of the brain may be analogous to trying to understand a modern computer by studying only its buses while ignoring its processors, memory, and software.

\subsection{Statistics: The Frequentist Lock-In}
\label{sec:case-stat}

The null-hypothesis significance testing (NHST) framework, based on the work of Fisher, Neyman, and Pearson in the early 20th century, became the dominant statistical methodology in science largely for contingent historical reasons: it was computationally tractable in the pre-computer era, it provided a simple decision procedure (reject or fail to reject), and it was aggressively promoted through influential textbooks.

The consequences have been severe. The replication crisis in psychology, medicine, and other fields is substantially attributable to the misuse and misunderstanding of $p$-values--a misunderstanding that is, arguably, \emph{invited} by the framework itself. The American Statistical Association's 2016 statement on $p$-values \citep{wasserstein2016asa} catalogued the ways in which standard practice diverges from the mathematical foundations.

Alternative frameworks offer substantive improvements:

\medskip\noindent\textit{Bayesian inference.} By directly computing posterior probabilities of hypotheses given data, Bayesian methods avoid many of the interpretive pitfalls of NHST. They handle nuisance parameters, model comparison, and sequential updating naturally. Their historical disadvantage--computational intractability--has been eliminated by modern MCMC methods.

\medskip\noindent\textit{Information-theoretic model selection.} The Akaike Information Criterion (AIC) and related methods evaluate models based on predictive accuracy rather than null-hypothesis rejection, avoiding the arbitrary dichotomy of ``significant'' vs.\ ``non-significant.''

\medskip\noindent\textit{Causal inference.} The frameworks of Pearl \citep{pearl2009causality} and Rubin provide principled methods for distinguishing correlation from causation--a distinction that NHST notoriously obscures.

The persistence of NHST despite these superior alternatives is perhaps the clearest example of institutional lock-in in all of science. Journals require $p$-values; textbooks teach NHST; reviewers expect significance tests. The switching cost is borne by the entire scientific community simultaneously, creating a collective action problem.

\subsection{Taxonomy: Trees vs.\ Networks}
\label{sec:case-tax}

Linnaeus's hierarchical classification of life--kingdom, phylum, class, order, family, genus, species--imposes a tree structure on biological diversity. Phylogenetics has refined this into evolutionary trees (cladograms) based on shared ancestry. But the tree of life is, in important respects, not a tree.

Horizontal gene transfer--the movement of genetic material between organisms outside of parent-to-offspring inheritance--is pervasive in prokaryotes and increasingly recognized in eukaryotes. Endosymbiosis (the origin of mitochondria and chloroplasts) represents a wholesale merger of lineages. Hybridization between species is common in plants and not rare in animals. Viral integration has contributed substantially to mammalian genomes.

These processes create a \emph{web} or \emph{network} of life, not a tree. Yet the institutional and conceptual infrastructure of biology--species concepts, taxonomic databases, phylogenetic software--is overwhelmingly tree-based. Network-based phylogenetic methods exist but remain marginal.

\subsection{Thermodynamics: The Equilibrium Bias}
\label{sec:case-thermo}

Classical thermodynamics was developed to describe heat engines--systems near thermal equilibrium undergoing quasi-static processes. Its central concepts (temperature, entropy, free energy) are well-defined only at or near equilibrium. The extension to non-equilibrium systems has been extraordinarily difficult, and much of the conceptual vocabulary of thermodynamics becomes ambiguous or undefined far from equilibrium.

Yet the most interesting systems in nature--living organisms, planetary atmospheres, economies, ecosystems--are far-from-equilibrium dissipative structures maintained by continuous energy flows. Prigogine's non-equilibrium thermodynamics \citep{prigogine1977self} and Jaynes's maximum entropy formalism \citep{jaynes1957information} represent attempts to extend thermodynamic reasoning beyond equilibrium, but neither has achieved the universal applicability of the equilibrium theory.

The possibility remains that an entirely different conceptual framework--one not descended from the steam-engine physics of Carnot and Clausius--might describe far-from-equilibrium systems more naturally. Information-theoretic approaches, viewing thermodynamics as a theory of inference rather than a theory of heat \citep{jaynes1957information}, represent one such candidate.

\section{Objections and Counterarguments}
\label{sec:counterarguments}

Our thesis faces several serious objections, which we address in turn.

\subsection{The Convergence Objection}

\textit{Objection:} Science does escape local minima. The history of science is punctuated by revolutions--Copernican astronomy, Darwinian evolution, relativity, quantum mechanics--that represent precisely the discontinuous jumps our thesis claims are missing.

\textit{Response:} We do not deny that paradigm shifts occur. Our claim is that they are \emph{insufficiently frequent and insufficiently radical} relative to the size of the landscape. Each revolution, moreover, tends to preserve much of the formal infrastructure of its predecessor. Quantum mechanics abandoned determinism but preserved differential equations; relativity abandoned absolute space and time but preserved the calculus of smooth manifolds. These are jumps within a relatively small region of framework space, not explorations of fundamentally alien formalisms.

\subsection{The Empirical Constraint Objection}

\textit{Objection:} Science is constrained by empirical data. Whatever formalism we use, it must agree with observation. This constraint drastically reduces the landscape, and within the reduced landscape, our current frameworks may be near-optimal.

\textit{Response:} Empirical adequacy constrains the \emph{predictions} of a framework, not its form. Many different formalisms can be empirically equivalent while differing enormously in explanatory power, computational tractability, and capacity for generalization. Lagrangian and Newtonian mechanics make identical predictions but differ profoundly in their capacity for extension (Lagrangian mechanics generalizes naturally to field theory and general relativity; Newtonian mechanics does not). The existence of empirically equivalent but structurally distinct frameworks is precisely what makes the landscape rugged.

\subsection{The Optimality Objection}

\textit{Objection:} Our current frameworks \emph{are} optimal, given human cognitive limitations. The globally optimal framework might be incomprehensible to human minds, in which case pursuing it is futile.

\textit{Response:} This objection has force, but it proves less than it claims. First, the constraint of human comprehensibility is itself changing: computational tools increasingly allow us to work with frameworks that exceed unaided human cognition. Second, some apparently ``intrinsic'' cognitive limitations may be products of training rather than biology. Mathematicians who grow up with category theory find it intuitive in ways that their set-theory-trained predecessors did not. Third, even if human comprehensibility is a hard constraint, the accessible region of framework space is still large enough to contain multiple local optima.

\subsection{The No-Free-Lunch Objection}

\textit{Objection:} No-free-lunch theorems in optimization show that no search strategy is universally superior. Our current approach may be locally suboptimal for some problems but globally efficient across all problems.

\textit{Response:} No-free-lunch theorems apply to optimization over \emph{arbitrary} landscapes. Real scientific landscapes are not arbitrary; they have structure imposed by the regularity of natural law. In structured landscapes, some search strategies are genuinely superior to others, and there is no reason to believe that historical gradient descent is among them.

\section{Strategies for Escaping Local Optima}
\label{sec:escape}

If our thesis is correct, what can be done? We propose several meta-scientific strategies, each inspired by techniques used to escape local minima in optimization.

\subsection{Simulated Annealing: Introducing Controlled Randomness}

In optimization, simulated annealing escapes local minima by occasionally accepting moves that \emph{increase} the cost function, with the probability of acceptance decreasing over time. The scientific analogue is the deliberate funding of high-risk, paradigm-challenging research--even when it does not yet demonstrate clear superiority over existing approaches.

Concrete proposals include: dedicated funding streams for ``anti-paradigmatic'' research (explicitly rewarding work that challenges foundational assumptions); sabbatical programs that encourage scientists to learn formalisms entirely outside their training; and ``red team'' review panels that evaluate the foundational assumptions of successful research programs.

\subsection{The Return to First Principles: Going Back to the Fork in the Road}
\label{sec:return}

Perhaps the most powerful strategy for escaping a local minimum is not to push forward but to go \emph{back}--to return to the historical moment where a critical choice was made and explore the path not taken. This strategy, which we might call \emph{principled regression}, involves identifying the foundational assumptions of a field, tracing them to their historical origins, and asking whether alternatives available at the time--but not pursued--might yield superior frameworks.

\medskip\noindent\textit{Taha's variational theory of lift: a paradigm case.} The most striking contemporary example of this strategy is the work of Haithem Taha and his group at the University of California, Irvine, who developed a new theory of aerodynamic lift by returning to first principles in classical mechanics that were available--but never applied--to the pioneers of aviation \citep{taha2022variational}.

The standard theory of lift, taught in every aeronautical engineering program worldwide, is attributed to Martin Kutta (1910). Kutta's theory requires that the airfoil have a single sharp trailing edge and relies on the \emph{Kutta condition}--a closure condition that selects a unique solution from the infinitely many solutions to Euler's equation for flow over a body. The Kutta condition is typically justified on the grounds that viscous effects (as described by the Navier-Stokes equations) enforce smooth flow at the trailing edge. This framework has been the backbone of aerodynamics for over a century.

Yet Kutta's theory fails for wings with multiple sharp edges, no sharp edges, or in unsteady flows. These are not exotic edge cases--they include insect wings, unconventional aircraft designs, and virtually all real-world unsteady aerodynamic phenomena.

Taha's breakthrough came not from developing more sophisticated computational tools or more powerful approximations to the Navier-Stokes equations, but from stepping outside the Newtonian framework entirely. He applied the \emph{Hertz principle of least curvature}--a variational principle in analytical mechanics that is rarely found in modern physics textbooks but was available in the nineteenth century--to develop a variational analogue of Euler's equations for ideal fluid dynamics. The resulting theory derives lift from a minimization principle: nature selects the flow configuration that minimizes a fundamental quantity (the Appellian, or the ``curvature'' of the system's trajectory in configuration space). The Kutta condition emerges as a special case for sharp-edged airfoils, reinterpreted not as a viscous effect but as a consequence of conservation of momentum.

The implications are profound. As Taha noted, the intellectual tools needed for this discovery were available to Kutta, Zhukovsky, Prandtl, and Von K\'{a}rm\'{a}n themselves--the early pioneers simply chose a different path. The puzzle was solved not by advancing along the existing gradient but by returning to the fork in the road and taking the other branch. Taha himself spent years studying the history and philosophy of mechanics--knowledge deemed ``useless'' by contemporary metrics--before recognizing the opportunity. This is precisely the kind of lateral exploration that our thesis identifies as essential for escaping local optima.

Moreover, Taha's group extended this variational approach to propose an entirely new school of fluid mechanics based on the \emph{principle of minimum pressure gradient} (PMPG) \citep{taha2023pmpg}, which offers an alternative to the Navier-Stokes paradigm for a range of problems. They demonstrated simple analytical solutions for flow over rotating cylinders--solutions that had never been found through the standard Navier-Stokes approach because the nonlinear PDE was too difficult to solve directly.

\medskip\noindent\textit{Hestenes and the geometric algebra revival.} A parallel example is David Hestenes's revival of geometric algebra (Clifford algebra) as a unified language for physics, beginning in the 1960s \citep{hestenes1984clifford}. William Kingdon Clifford had developed this algebraic framework in 1878, unifying Grassmann's exterior algebra with Hamilton's quaternions into a single coherent system. But Clifford died young, and the rival framework of Gibbs's vector calculus--inferior in generality but simpler to teach--won the pedagogical competition. For nearly a century, geometric algebra languished in obscurity while physicists juggled an awkward patchwork of vectors, tensors, spinors, differential forms, and matrix algebras--each capturing part of the geometric structure that Clifford's algebra handles uniformly.

Hestenes recognized that many of the apparent complexities in physics--the mysterious role of complex numbers in quantum mechanics, the awkwardness of cross products in three dimensions, the seeming incompatibility of different algebraic frameworks--were artifacts of the historical choice to adopt Gibbs's vector calculus over Clifford's geometric algebra. By returning to the nineteenth-century fork, he developed a framework in which Maxwell's equations reduce to a single equation, the Dirac equation acquires a transparent geometric meaning, and the proliferation of distinct algebraic systems is replaced by a single unified language. The framework has since found applications in computer graphics, robotics, and computational geometry--but it remains marginal in physics education, a testament to the power of institutional lock-in.

\medskip\noindent\textit{Constructor theory: rethinking the foundations of physics.} Chiara Marletto and David Deutsch's constructor theory represents another attempt to return to foundations. Rather than describing what \emph{does} happen (the trajectory of a system through state space, as in Newtonian mechanics), constructor theory describes what \emph{can} and \emph{cannot} happen--which transformations are possible and which are impossible. This reframing, inspired by foundational questions in quantum information theory, may dissolve problems that appear intractable in the standard framework by approaching them from a fundamentally different angle.

\medskip\noindent\textit{Lessons from principled regression.} These examples share a common structure: a researcher steps outside the active frontier of their field, studies its historical and philosophical foundations, identifies a critical juncture where an alternative path was available but not taken, and then develops that alternative with modern tools and insights. This requires a combination of deep historical knowledge, philosophical sophistication, and technical skill--precisely the combination that modern scientific training, with its emphasis on narrow specialization, tends to discourage.

The lesson for escaping local optima is clear: \emph{the most productive direction may be backward}. The foundational assumptions of a field, precisely because they are so deeply buried, are the most likely source of hidden constraints. And the history of science is littered with roads not taken--variational principles not applied, algebraic systems not developed, foundational frameworks not pursued--any of which might open up vast new regions of the landscape.

\subsection{Momentum Methods: Sustaining Radical Programs}

Momentum-based optimizers (like Adam or RMSProp in machine learning) allow the search to continue moving in a productive direction even when the local gradient becomes shallow. The scientific analogue is providing long-term, stable support for radical research programs that have not yet produced results. The current funding model, which demands frequent demonstrations of progress, is poorly suited to paradigm-challenging work that may require decades of development before yielding fruit.

Taha's case is again instructive: he spent years studying the history and philosophy of mechanics before the key insight emerged, a period during which no publications resulted. Under the standard metrics of academic productivity, this would register as wasted time. Only a combination of personal conviction and institutional tolerance allowed the work to continue long enough to produce its breakthrough.

\subsection{Population-Based Methods: Maintaining Diversity}

Evolutionary algorithms maintain a \emph{population} of candidate solutions, exploring multiple regions of the landscape simultaneously. The scientific analogue is deliberately maintaining diverse research programs and resisting the tendency toward methodological monoculture. This requires active intervention, since competition for resources naturally drives convergence.

Concretely, funding agencies should allocate a fixed percentage of their budget to programs that \emph{explicitly} challenge the foundational assumptions of dominant paradigms. Journals should create tracks for ``foundational alternatives'' that are reviewed by criteria different from incremental contributions. And universities should create positions for scholars whose primary contribution is the development of new frameworks, not the solution of problems within existing ones.

\subsection{Transfer Learning: Cross-Disciplinary Pollination}

Some of the most productive paradigm shifts in science have occurred when methods from one field were imported into another: Shannon's application of thermodynamic concepts to communication, the application of gauge theory from physics to machine learning, the use of category theory in computer science. Systematic programs for cross-disciplinary training and collaboration--going beyond superficial ``interdisciplinarity'' to genuine immersion in alien formalisms--may be among the most effective strategies for landscape exploration.

\subsection{Artificial Intelligence as a Landscape Explorer}

Machine learning systems are, in a sense, formalisms that are not subject to human cognitive biases. The success of neural networks in discovering novel strategies in games (AlphaGo's ``alien'' moves), in protein structure prediction (AlphaFold), and in theorem proving (Lean-based systems) suggests that AI may serve as a tool for exploring regions of framework space that human scientists would never reach. The framework of ``AI-driven science'' or ``automated science'' may ultimately be the most powerful strategy for escaping human cognitive lock-in. However, this optimistic assessment faces a fundamental objection that must be addressed directly.

\section{The Bootstrap Paradox: Can AI Escape What Created It?}
\label{sec:bootstrap}

Any serious proposal that artificial intelligence might help science escape local optima must confront what we call the \emph{bootstrap paradox}: AI systems are trained on the corpus of existing human knowledge--precisely the body of knowledge that constitutes the local minimum. If the training data is non-optimal, how can the output be anything other than a sophisticated recombination of non-optimal ideas? The AI has never seen the global minimum because no human has written about it.

This paradox has teeth, and we should not dismiss it lightly. We argue, however, that it is ultimately resolvable, and that the resolution illuminates important features of both AI and scientific progress.

\subsection{The Paradox Stated Formally}

Let $\mathcal{D}$ denote the training corpus of a large AI system--essentially the digitized record of human scientific knowledge. By our thesis, $\mathcal{D}$ is generated from within the basin of attraction of the local minimum $f_{\text{now}}$. An AI system $A$ trained on $\mathcal{D}$ learns a distribution $P_A$ over possible outputs that is, by construction, concentrated in the neighborhood of $f_{\text{now}}$.

The paradox: if $P_A$ is anchored to $f_{\text{now}}$, how can $A$ generate outputs in the vicinity of $f^*$, the global optimum, which by hypothesis lies in a different basin entirely? This would seem to require the AI to extrapolate beyond its training distribution--precisely the regime where modern AI systems are least reliable.

\subsection{Five Reasons the Paradox Is Not Fatal}

Despite its force, we identify five reasons why AI may still contribute to escaping local optima.

\medskip\noindent\textit{(i) The training data contains the roads not taken.} The corpus of human knowledge does not contain only the dominant paradigm. It also contains the historical record of abandoned alternatives, minority programs, and forgotten formalisms. Clifford's geometric algebra, Hertz's principle of least curvature, Lanchester's rejected lift theory, Bayesian statistics before the MCMC revolution, category theory before its current resurgence--all of these are \emph{in the training data}. They are simply not connected to the problems they might solve, because no human has yet made the connection. An AI system, with its capacity to hold the entire corpus simultaneously and detect statistical regularities across disconnected literatures, can serve as a \emph{cross-pollination engine}--not generating genuinely novel formalisms, but discovering unexploited connections between existing ones.

This is precisely what principled regression requires: matching forgotten tools to current problems. An AI can search the space of ``historical formalism $\times$ contemporary open problem'' far more exhaustively than any human scientist.

\medskip\noindent\textit{(ii) The training data encodes the fingerprints of failure.} The corpus contains not only the successes of the current paradigm but also its \emph{failures}: unsolved problems, anomalous experimental results, persistent discrepancies, the replication crisis data, papers that say ``we do not understand why this does not work.'' These failures are \emph{signatures of the basin walls}--they mark the directions in which the current framework strains and breaks. An AI system trained to recognize patterns of failure, rather than patterns of success, might map the contours of the local minimum even if it cannot see over the rim.

Consider: if an AI could identify that the cluster of difficulties in turbulence, protein folding, and consciousness share a common structural feature--say, they all involve systems where the Newtonian paradigm of differential equations encounters nonlinearities that resist linearization--this would be a powerful diagnostic, even if the AI cannot itself propose the alternative formalism that dissolves all three.

\medskip\noindent\textit{(iii) AI does not inherit the \emph{mechanisms} of lock-in.} A critical distinction must be drawn between the \emph{content} of human knowledge (which AI inherits) and the \emph{mechanisms} that maintain lock-in (which it does not). AI has no linearization bias, no spatial-visual bias, no reductionist cognitive template. It has no career to protect, no funding agency to please, no Ph.D.\ in a specific formalism that creates switching costs. It has no national identity that privileges one school of thought over another. It has no difficulty treating Bayesian and frequentist statistics with equal seriousness, or considering that the Navier-Stokes equations might not be the best framework for fluid dynamics.

This means that even when working within the training distribution, AI can \emph{weight} alternatives differently than humans do. An alternative that every human physicist knows about but dismisses for sociological reasons (``that approach went out of fashion in the 1970s'') might be treated seriously by an AI that has no sociological commitments. The content is the same; the evaluation function is different.

\medskip\noindent\textit{(iv) AI can explore formal structures without semantic anchoring.} Modern AI systems, particularly in mathematics, can manipulate formal structures--algebraic systems, logical frameworks, combinatorial objects--without the semantic commitments that constrain human reasoning. A human mathematician trained in set theory thinks \emph{in terms of} sets; a category theorist thinks \emph{in terms of} morphisms. These semantic anchors make some manipulations natural and others unthinkable. An AI system that treats formal structures as patterns to be extended, combined, and transformed may explore regions of formal space that are semantically inaccessible to human mathematicians--not because the AI is smarter, but because it is not anchored.

The success of AI in discovering novel mathematical conjectures \citep{davies2021advancing}--identifying patterns in mathematical data that human mathematicians had not noticed--provides early evidence for this capacity. These conjectures were not in the training data; they were latent structures that became visible only when the data was examined without the usual semantic filters.

\medskip\noindent\textit{(v) AI enables simulation of counterfactual scientific histories.} Perhaps the most speculative but intriguing possibility: AI might be used to simulate \emph{alternative histories of science}. What if we asked an AI system to develop physics starting from Leibniz's relational conception of space rather than Newton's absolute space? What if we asked it to build chemistry starting from the electron density rather than from molecular bonds? What if we asked it to construct biology starting from information-theoretic principles rather than gene-centric ones?

Each of these counterfactual exercises would, in effect, simulate a random jump in the framework landscape--a different starting point for the gradient descent of normal science. The results would not necessarily be superior to current science, but they would explore different basins of attraction. By comparing multiple counterfactual scientific histories, we might identify frameworks that consistently outperform alternatives across multiple starting points--a form of ensemble optimization over the landscape of possible sciences.

\subsection{The Deeper Risk: AI as Lock-In Amplifier}

We must also acknowledge the possibility that AI could \emph{deepen} rather than escape the local minimum. If AI systems are deployed primarily to accelerate research within existing paradigms--faster drug discovery using existing molecular frameworks, more efficient CFD simulations using existing numerical methods, better protein structure prediction within existing biochemical ontologies--they will make the current paradigm more productive and thereby increase the opportunity cost of abandoning it. The basin of attraction would deepen.

This is not a hypothetical concern. The overwhelming majority of ``AI for science'' research today is paradigm-reinforcing rather than paradigm-challenging. AI is being used to solve problems \emph{faster within} existing frameworks, not to question whether those frameworks are optimal. If this trend continues, AI may become the most powerful mechanism of lock-in ever created: an amplifier of the current gradient that makes the local minimum so comfortable that the incentive to explore the wider landscape vanishes entirely.

\subsection{Designing AI for Paradigm Exploration}

To avoid this trap, we propose that a deliberate effort be made to design AI systems specifically for paradigm exploration rather than paradigm acceleration:

\begin{enumerate}[label=(\roman*)]
  \item \textit{Cross-historical retrieval systems.} AI tools that systematically match forgotten or abandoned scientific frameworks with contemporary open problems, automating the process of principled regression that Taha performed manually.
  
  \item \textit{Anomaly clustering.} AI systems trained to identify structural similarities among the failures and anomalies of the current paradigm, mapping the basin walls rather than optimizing within the basin floor.
  
  \item \textit{Counterfactual science generators.} AI systems tasked with developing alternative scientific frameworks from different starting axioms, producing multiple candidate descriptions of the same phenomena for comparative evaluation.
  
  \item \textit{Formalism translators.} AI tools that can take a body of results expressed in one formalism (e.g., differential equations) and systematically re-express them in alternative formalisms (e.g., variational principles, algebraic structures, computational models), making it possible to identify which difficulties are intrinsic to the phenomena and which are artifacts of the language.
  
  \item \textit{Adversarial reviewers.} AI systems explicitly trained to challenge the foundational assumptions of research proposals and published work--scientific ``red teams'' that ask not ``Is this result correct within the current framework?'' but ``Is the current framework the right one for this problem?''
\end{enumerate}

The bootstrap paradox is real but not fatal. AI inherits the content of the local minimum but not the mechanisms that maintain it. If deployed wisely--as a tool for cross-pollination, anomaly detection, and counterfactual exploration rather than mere acceleration--AI may prove to be the most powerful instrument yet devised for exploring the wider landscape of possible sciences.

\section{Epistemological Implications}
\label{sec:implications}

Our thesis carries deep implications for the philosophy of science.

\subsection{Against Scientific Triumphalism}

If current scientific knowledge is a local optimum rather than a global one, then the widespread attitude of scientific triumphalism--the belief that our current theories are approximately final--is unwarranted. This does not imply scientific skepticism: our theories are empirically successful and represent genuine knowledge about nature. But their form--the formalisms, the conceptual vocabularies, the methodological principles--may be substantially improvable.

\subsection{Revisiting Kuhn and Lakatos}

Our thesis extends Kuhn's theory of scientific revolutions \citep{kuhn1962structure} by providing a formal model of why revolutions are both necessary and rare: the rugged landscape explains why normal science converges to local optima (paradigm consolidation) and why escape requires discontinuous jumps (revolutions). It also explains Lakatos's observation \citep{lakatos1978methodology} that degenerating research programs can persist long after superior alternatives are available: the switching cost (the energy barrier in the landscape) can exceed the improvement offered by the new paradigm.

\subsection{The Contingency of Science}

Perhaps the most profound implication is that science is more contingent than commonly supposed. An alien civilization, starting with different cognitive architectures, different mathematical traditions, and different initial problems, might develop a body of scientific knowledge that is empirically equivalent to ours but formally unrecognizable. They might find our differential equations quaint, our concept of a chemical bond arbitrary, and our gene-centric biology provincial--not because their science is better, but because they found a different local minimum that happens to be deeper than ours.

This possibility should inspire not despair but ambition. The scientific landscape is vast, and we have explored only a tiny corner of it. The tools for exploring further--computational methods, artificial intelligence, cross-disciplinary synthesis, and deliberate institutional reform--are increasingly available. The question is whether we have the intellectual courage to use them.

\section{Conclusion}
\label{sec:conclusion}

We have argued that the current body of scientific knowledge represents a local optimum in a vast landscape of possible frameworks, and that this non-optimality is maintained by four interlocking mechanisms: cognitive lock-in (the biases of human thought), formal lock-in (the inertia of mathematical and notational infrastructure), institutional lock-in (the feedback loops of funding, publication, and prestige), and sociopolitical lock-in (the channeling of science by wars, geopolitical rivalries, and colonial power structures). We have supported this thesis with case studies drawn from across the sciences, each illustrating how the perceived difficulty of open problems may partly reflect the limitations of our frameworks rather than the intrinsic complexity of nature.

Crucially, we have identified a concrete strategy for escaping local optima that is already producing results: \emph{principled regression}--the deliberate return to historical junctures where alternative paths were available but not taken. The work of Taha on variational aerodynamics, of Hestenes on geometric algebra, and of others who have gone back to the basics to find what was missed, demonstrates that the most important breakthroughs may come not from pushing harder along the current gradient but from retracing our steps to explore abandoned forks in the road.

We have also confronted the bootstrap paradox of artificial intelligence: AI systems trained on the products of our local minimum inherit its content but not the cognitive, sociological, and political mechanisms that maintain it. This distinction is the key to understanding how AI can contribute to paradigm exploration rather than merely paradigm acceleration. If deployed wisely--as cross-pollination engines, anomaly detectors, and counterfactual science generators--AI may prove to be the most powerful instrument ever devised for exploring the wider landscape. If deployed naively, as a tool for faster optimization within existing frameworks, it may become the most powerful mechanism of lock-in ever created.

Our thesis is not a counsel of despair. It is, rather, an invitation to take seriously the possibility that the most important scientific breakthroughs of the future may come not from solving hard problems within existing frameworks, but from discovering that different frameworks dissolve those problems entirely. The gradient descent of normal science has brought us far. The time has come to explore the wider landscape.

\vspace{1cm}

\subsection*{Acknowledgments}
[To be added.]



\begin{thebibliography}{99}

\bibitem[Bader(1990)]{bader1990atoms}
Bader, R.~F.~W. (1990).
\newblock \emph{Atoms in Molecules: A Quantum Theory}.
\newblock Oxford University Press.

\bibitem[Beniaguev et~al.(2021)]{beniaguev2021single}
Beniaguev, D., Segev, I., \& London, M. (2021).
\newblock Single cortical neurons as deep artificial neural networks.
\newblock \emph{Neuron}, 109(17), 2727--2739.

\bibitem[Bloor(2011)]{bloor2011enigma}
Bloor, D. (2011).
\newblock \emph{The Enigma of the Aerofoil: Rival Theories in Aerodynamics, 1909--1930}.
\newblock University of Chicago Press.

\bibitem[Crutchfield(2012)]{crutchfield2012between}
Crutchfield, J.~P. (2012).
\newblock Between order and chaos.
\newblock \emph{Nature Physics}, 8, 17--24.

\bibitem[Davies et~al.(2021)]{davies2021advancing}
Davies, A., Veli\v{c}kovi\'{c}, P., Buesing, L., et~al. (2021).
\newblock Advancing mathematics by guiding human intuition with {AI}.
\newblock \emph{Nature}, 600(7887), 70--74.

\bibitem[Fields(2009)]{fields2009other}
Fields, R.~D. (2009).
\newblock \emph{The Other Brain: The Scientific and Medical Breakthroughs That Will Heal Our Brains and Revolutionize Our Health}.
\newblock Simon \& Schuster.

\bibitem[Frisch et~al.(1986)]{frisch1986lattice}
Frisch, U., Hasslacher, B., \& Pomeau, Y. (1986).
\newblock Lattice-gas automata for the {N}avier-{S}tokes equation.
\newblock \emph{Physical Review Letters}, 56(14), 1505--1508.

\bibitem[Jaynes(1957)]{jaynes1957information}
Jaynes, E.~T. (1957).
\newblock Information theory and statistical mechanics.
\newblock \emph{Physical Review}, 106(4), 620--630.

\bibitem[Kauffman(1993)]{kauffman1993origins}
Kauffman, S.~A. (1993).
\newblock \emph{The Origins of Order: Self-Organization and Selection in Evolution}.
\newblock Oxford University Press.

\bibitem[Kuhn(1962)]{kuhn1962structure}
Kuhn, T.~S. (1962).
\newblock \emph{The Structure of Scientific Revolutions}.
\newblock University of Chicago Press.

\bibitem[Lakatos(1978)]{lakatos1978methodology}
Lakatos, I. (1978).
\newblock \emph{The Methodology of Scientific Research Programmes: Philosophical Papers Volume 1}.
\newblock Cambridge University Press.

\bibitem[Maldacena(1999)]{maldacena1999large}
Maldacena, J. (1999).
\newblock The large-{$N$} limit of superconformal field theories and supergravity.
\newblock \emph{International Journal of Theoretical Physics}, 38(4), 1113--1133.

\bibitem[McFadden(2020)]{mcfadden2020integrating}
McFadden, J. (2020).
\newblock Integrating information in the brain's {EM} field: the cemi field theory of consciousness.
\newblock \emph{Neuroscience of Consciousness}, 2020(1), niaa016.

\bibitem[Noble(2006)]{noble2006music}
Noble, D. (2006).
\newblock \emph{The Music of Life: Biology Beyond Genes}.
\newblock Oxford University Press.

\bibitem[Pearl(2009)]{pearl2009causality}
Pearl, J. (2009).
\newblock \emph{Causality: Models, Reasoning, and Inference} (2nd ed.).
\newblock Cambridge University Press.

\bibitem[Nicolis \& Prigogine(1977)]{prigogine1977self}
Nicolis, G. \& Prigogine, I. (1977).
\newblock \emph{Self-Organization in Nonequilibrium Systems: From Dissipative Structures to Order through Fluctuations}.
\newblock Wiley.

\bibitem[Succi(2018)]{succi2018lattice}
Succi, S. (2018).
\newblock \emph{The Lattice Boltzmann Equation: For Complex States of Flowing Matter}.
\newblock Oxford University Press.

\bibitem[Taha \& Gonzalez(2022)]{taha2022variational}
Taha, H.~E. \& Gonzalez, C. (2022).
\newblock A variational theory of lift.
\newblock \emph{Journal of Fluid Mechanics}, 941, A58.

\bibitem[Taha et~al.(2023)]{taha2023pmpg}
Taha, H.~E., Gonzalez, C., \& Shorbagy, M. (2023).
\newblock A minimization principle for incompressible fluid mechanics.
\newblock \emph{Physics of Fluids}, 35(12).

\bibitem[Hestenes \& Sobczyk(1984)]{hestenes1984clifford}
Hestenes, D. \& Sobczyk, G. (1984).
\newblock \emph{Clifford Algebra to Geometric Calculus: A Unified Language for Mathematics and Physics}.
\newblock D.~Reidel Publishing Company.

\bibitem[Wasserstein \& Lazar(2016)]{wasserstein2016asa}
Wasserstein, R.~L. \& Lazar, N.~A. (2016).
\newblock The {ASA} statement on $p$-values: context, process, and purpose.
\newblock \emph{The American Statistician}, 70(2), 129--133.

\end{thebibliography}
\end{document}